\documentclass[conference]{IEEEtran}
\IEEEoverridecommandlockouts
\usepackage{cite}
\usepackage{amsmath,amssymb,amsfonts}
\usepackage{algorithmic}
\usepackage{graphicx}
\usepackage{algorithm}
\usepackage{subcaption}
\usepackage{textcomp}
\usepackage{multirow}
\usepackage{comment}
\usepackage{xcolor}

\definecolor{mycustompurple}{RGB}{154, 36, 79} 
\usepackage[utf8]{inputenc}
\usepackage{hyperref} 

\hypersetup{
    colorlinks=true,            
    linkcolor=red,              
    citecolor=green,            
    filecolor=mycustompurple,   
    urlcolor=magenta            
}

\def\BibTeX{{\rm B\kern-.05em{\sc i\kern-.025em b}\kern-.08em
    T\kern-.1667em\lower.7ex\hbox{E}\kern-.125emX}}
\begin{document}

\title{LEA: Label Enumeration Attack in Vertical Federated Learning}

\author{
    \IEEEauthorblockN{Wenhao Jiang, Shaojing Fu\IEEEauthorrefmark{2}\thanks{\IEEEauthorrefmark{2}The corresponding author.}, Yuchuan Luo, Lin Liu}
    \IEEEauthorblockA{College of Computer Science and Technology, National University of Defense Technology, Changsha, China}
    \IEEEauthorblockA{\{jwh\_roy, fushaojing, luoyuchuan09, liulin16\}@nudt.edu.cn}
}

\maketitle
\begin{abstract}

A typical Vertical Federated Learning (VFL) scenario involves several participants collaboratively training a machine learning model, where each party has different features for the same samples, with labels held exclusively by one party. Since labels contain sensitive information, VFL must ensure the privacy of labels. However, existing VFL-targeted label inference attacks are either limited to specific scenarios or require auxiliary data, rendering them impractical in real-world applications.

We introduce a novel Label Enumeration Attack (LEA) that, for the first time, achieves applicability across multiple VFL scenarios and eschews the need for auxiliary data. Our intuition is that an adversary, employing clustering to enumerate mappings between samples and labels, ascertains the accurate label mappings by evaluating the similarity between the benign model and the simulated models trained under each mapping. To achieve that, the first challenge is how to measure model similarity, as models trained on the same data can have different weights. Drawing from our findings, we propose an efficient approach for assessing congruence based on the cosine similarity of the first-round loss gradients, which offers superior efficiency and precision compared to the comparison of parameter similarities. However, the computational cost may be prohibitive due to the necessity of training and comparing the vast number of simulated models generated through enumeration. To overcome this challenge, we propose Binary-LEA from the perspective of reducing the number of models and eliminating futile training, which lowers the number of enumerations from $\mathcal{O}(n!)$ to $\mathcal{O}(n^3)$. Experiments across different VFL settings confirm LEA's effectiveness. In the absence of auxiliary datasets, LEA demonstrates a 50\% to 90\% enhancement in attack accuracy over the existing state-of-the-art label inference attacks in VFL. Moreover, LEA is resilient against common defense mechanisms such as gradient noise and gradient compression. 

\end{abstract}


\section{Introduction}
Vertical Federated Learning (VFL) is an emerging paradigm in the field of machine learning that addresses the challenges of data privacy and centralization. Unlike traditional Federated Learning (FL), which operates on a single, albeit large, dataset spread across multiple clients \cite{wu2025efficient}, VFL specifically deals with scenarios where each clients hold different but complementary feature sets of the same sample space. For instance, WeBank employs VFL to collaborate with invoice agencies to establish financial risk models for their corporate clients \cite{cheng2020federated}. In recent years, the demand for VFL has emerged and grown robustly within the industry. Companies and institutions with limited and fragmented data have been seeking to compensate by partnering with data collaborators to jointly develop artificial intelligence technologies, aiming to maximize the utility of their data \cite{li2020review,li2021survey}. This collaborative approach allows for the training of robust machine learning models without the need to pool data from various sources, thus respecting data privacy and ownership constraints. 

In VFL, each party contributes to the learning process by providing his unique features, and a central server coordinates the training algorithm without having access to the raw data. In a typical VFL scenario, only one party possesses the label information of the dataset, which is called as the \textbf{active party}. The remaining participants, who do not have access to the labels, are referred to as \textbf{passive parties}. 

VFL is considered secure because only intermediate values are exchanged between the active party and the passive parties, which is believed to prevent the disclosure of the passive parties' feature information and the active party's label information \cite{li2021survey}. However, Fu et al \cite{fu2022label} proposed label inference attacks against VFL, where the passive party can generate a complete model to infer the labels with access to a small portion of the dataset. They have introduced, for the first time, a label inference attack within VFL, which has catalyzed further research into privacy threats and defenses within the VFL domain. This form of attack is often targeted by defensive solutions \cite{li2023fedvs,lu2023squirrel} and can be considered a cornerstone in the field of VFL privacy and security. 

However, the effectiveness of this attack is highly dependent on the quality of the dataset acquired by the adversary. This exposes the limitation of existing work, which is the requirement for auxiliary data to achieve good attack effectiveness. In this work, we discovered that even without access to a subset of the dataset, our method enables the passive party to carry out a label inference attack, revealing privacy and security threats inherent in VFL. The \textbf{Label Enumeration Attack (LEA)} we propose is similar to brute-force cryptanalysis, enumerating over possible label permutations to train simulated models. By comparing the similarity between these simulated models and a normally trained model, we get the attack model. It is assumed that the passive party's local data is inherently categorizable, a hypothesis we have confirmed through our experiments. This assumption is plausible in many scenarios, such as when the active party holds the labels and the passive party possesses the data, or when certain features of the passive party are particularly influential. However, without label information, the passive party cannot ascertain the correspondence between the samples and the true labels. Our work focuses on developing methods to uncover this correspondence. This research contributes to a deeper understanding of the privacy risks in VFL and provides insights into potential countermeasures to safeguard against such label inference attacks.

During the attack process of LEA, we encountered two primary challenges. The first is determining model similarity. For neural network models, even when two identically initialized models are trained on the same dataset, the final parameter similarity can be vastly different post-convergence. In the SplitVFL \cite{liu2024vertical} scenario, comparing the similarity of the bottom models presents a greater challenge than comparing the complete models. To address this, we propose assessing model similarity based on the first-round gradient differences under certain conditions. The second is reducing computational cost. For a multi-class classification task with $n$ labels, generating $n!$ label permutations corresponds to creating and training $n!$ simulated models, which is computationally expensive. To mitigate this, we propose the Binary-Label Enumeration Attack (Binary-LEA), which transforms the multi-class task into multiple binary classification tasks, thereby reducing the computational overhead to $\mathcal{O}(n^3)$.

Additionally, we propose a defense strategy and evaluate two commonly used VFL defense schemes: gradient noise \cite{zhu2019deep} and gradient compression \cite{kairouz2021advances}. We found that the proposed defense strategy based on a label mapping table can alleviate the effects of LEA to some extent, while the latter two typical label inference attack defense schemes perform poorly against LEA. This encourages subsequent research to delve deeper into the exploration of defense schemes in VFL.

The primary contributions of this paper can be summarized as follows:

\begin{itemize}
    \item We propose a novel Label Enumeration Attack (LEA) based on clustering algorithms. Compared to previous work, LEA is more suitable for situations without auxiliary datasets and various VFL scene settings.
    
    \item For model similarity assessment, we introduce a method based on the first-round gradient differences in specific conditions. To reduce computational cost, we propose Binary-LEA that converts multi-class tasks into multiple binary tasks, which lowers the number of enumerations from $\mathcal{O}(n!)$ to $\mathcal{O}(n^3)$.
    \item We evaluate our attack using real-world datasets in both two-party and multi-party settings, achieving significant performance. Additionally, we provide direct and rigorous theoretical evidence to confirm the accuracy of our attack. Furthermore, we assess two classic defensive methods and find them to be ineffective against our attack and propose a defense scheme that can alleviate LEA.
\end{itemize}

\section{Background and Related Work}

\subsection{Vertical Federated Learning}
In recent years, VFL \cite{Yongxin2019Federated} has received significant attention from both the academic community and industry, with hundreds of research articles being published annually. There has also been a proliferation of open-source projects, including notable contributions such as FATE \cite{liu2021fate}, Fedlearner, PaddleFL, PySyft \cite{ryffel2018generic}, FedTree \cite{li2022fedtree}, and FedML \cite{he2020fedml}. Moreover, real-world industrial applications of VFL have emerged in various sectors, including advertising \cite{cai2020bytedance} and finance \cite{cheng2020federated}.

\begin{figure}[htbp]
  \centering
  \includegraphics[width=\linewidth]{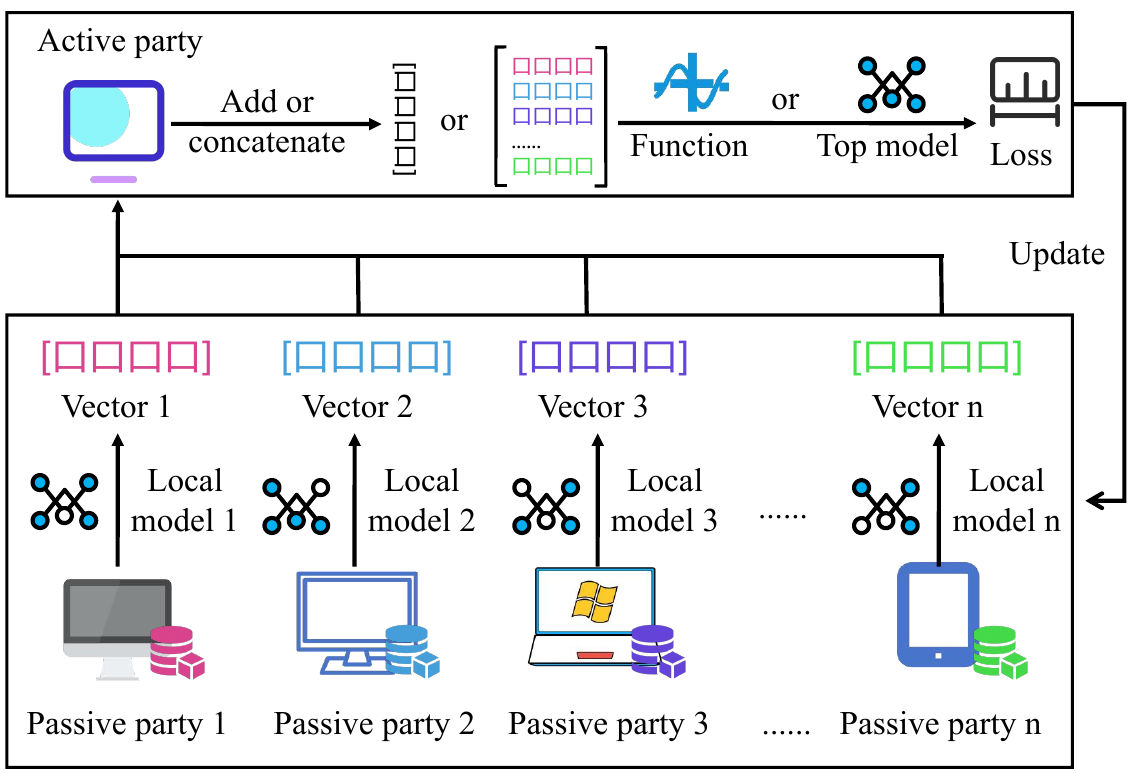}
  \caption{Flowchart of Vertical Federated Learning.}\label{VFLflow}
\end{figure}

VFL is designed for scenarios where participant datasets share the same sample space but have distinct feature spaces. In such a scenario, the features of the model are distributed among them, but the labels are proprietary to only one. The current mainstream perspective categorizes VFL into \textbf{AggVFL} \cite{zou2023vflair} and \textbf{SplitVFL} \cite{liu2024vertical} based on whether the active party's global model is trainable. When the global model acts as an aggregation function, typically the cross-entropy loss function, the scenario is known as AggVFL. On the other hand, if the global model is a trainable part of the model, the scenario is termed SplitVFL. We provide a flow diagram of the VFL process in Figure~\ref{VFLflow}.

\begin{table*}[htbp]
\renewcommand{\arraystretch}{1}
\caption{Label Inference Attacks in VFL.}
\begin{center}
\addtolength{\tabcolsep}{-3pt}
\begin{tabular}{cccccccccc}
\hline

\multirow{3}{*}{\textbf{Attack Scheme}}&\multicolumn{2}{c}{\multirow{2}{*}{\textbf{VFL Setting}}}& \multicolumn{2}{c}{\multirow{2}{*}{\textbf{Model}}}&\textbf{Support} & \textbf{Protecting} &\multirow{2}{*}{\textbf{Auxiliary}} & \textbf{Number} &\multirow{2}{*}{\textbf{Computational}} \\
& & & & &\textbf{Batch-Level}&\textbf{Intermediate} & &\textbf{of classes} \\
\cline{2-5} 
&\textbf{AggVFL}&\textbf{SplitVFL}&\textbf{LR}&\textbf{NN}&\textbf{Gradients}&\textbf{Results}&\textbf{Requirement} &\textbf{Supported} &\textbf{complexity$^{\mathrm{1}}$} \\
\hline
Direct Label Inference (DLI) \cite{fu2022label,li2021label}& $\checkmark$ &$\times$ & $\times$ & $\checkmark$ & $\times$ & $\times$ &- &$\geq 2$ & $\mathcal{O}(m)$ \\
Norm Scoring Attack (NS) \cite{li2021label}& $\times$& $\checkmark$ & $\times$& $\checkmark$ &$\times$ & $\times$&- &$=2$ & $\mathcal{O}(m)$  \\
Direction Scoring Attack (DS) \cite{li2021label}& $\times$& $\checkmark$ & $\times$& $\checkmark$&$\times$ & $\times$ &- &$=2$ & $\mathcal{O}(m)$  \\
Residual Reconstruction Attacks (RR) \cite{tan2022residue}& $\checkmark$ &$\times$ & $\checkmark$ &$\times$& $\checkmark$ & $\checkmark$&-&$\geq 2$ & $\mathcal{O}(d^3)$  \\
Passive Model Completion (PMC) \cite{fu2022label}& $\times$& $\checkmark$ & $\times$& $\checkmark$ & $\checkmark$ & $\checkmark$&Labeled data&$\geq 2$ & $\mathcal{O}(kTd+md)$  \\
Active Model Completion (AMC) \cite{fu2022label}& $\times$& $\checkmark$ & $\times$& $\checkmark$ & $\checkmark$ & $\checkmark$&Labeled data&$\geq 2$ & $\mathcal{O}(mTd)$  \\
Label-related Relation Inference (LRI) \cite{qiu2022your}& $\times$& $\checkmark$ & $\times$& $\checkmark$& $\checkmark$ & $\checkmark$&- &$=2$ & $\mathcal{O}(Td+md)$  \\
Gradient Inversion with a Label Prior \cite{kariyappa2023exploit}& $\times$& $\checkmark$ & $\times$& $\checkmark$& $\checkmark$& $\checkmark$& Label distribution &$=2$ & $\mathcal{O}(mTd)$  \\
Label Enumeration Attack (LEA, \textcolor{red}{this work}) & $\checkmark$ & $\checkmark$ & $\checkmark$& $\checkmark$ & $\checkmark$ & $\checkmark$&- &$\geq 2$ & $\mathcal{O}(n^3Td)$ \\
\hline
\multicolumn{10}{l}{$^{\mathrm{1}}$Here, $m$ denotes the number of samples, $n$ denotes the number of labels, $T$ denotes the number of training rounds, $d$ denotes the dimension of model} \\
\multicolumn{10}{l}{parameters, and $k$ denotes the number of samples in the auxiliary dataset.} \\
\end{tabular}
\label{tab:attacks}
\end{center}
\end{table*}

\subsection{Label Inference Attacks In VFL}

In the real world, labels possessed by the active party, such as patients' diagnostic results and personal loan default records, are considered sensitive information that should only be accessible to authorized entities. The privacy and security of label information are not only a prerequisite for the proper functioning of VFL but also a target for malicious passive parties. An adversary may exploit the training process in an attempt to infer the valuable labels held by the active party. This adversary could operate under the honest-but-curious security assumption, where they follow the protocol but attempt to glean information, or they could actively manipulate the protocol under a malicious assumption. Up to now, researchers have proposed various label inference attacks across different protocols.

When the training protocol employs \textbf{sample-level} gradients, the adversary can obtain sample-level gradients $\frac{\partial L}{\partial \hat{y}}$ propagated back from the active party where $L$ denotes loss function, $\hat{y}$ means outputs of passive party. They can utilize this information to conduct Direct Label Inference (DLI) \cite{fu2022label,li2021label}. For AggVFL, the active party employs non-trainable global model such as softmax, and DLI can achieve an attack with up to 100\% accuracy, because the gradient vector for each sample has only one element with the opposite sign to all other elements, thereby revealing the label \cite{li2021label}. For specific scenarios such as binary classification, even in SplitVFL, attackers can infer labels from sample-level gradients through norm scoring (NS) or direction scoring (DS) attacks \cite{li2021label}.

When the training protocol uses \textbf{batch-level} gradients, the adversary cannot obtain the sample-level gradients $\frac{\partial L}{\partial \hat{y}_s}$, but it might be able to obtain batch-level gradients $\frac{\partial L}{\partial \hat{y}_b}$ from the active party. Research indicates that in this scenario, it is still possible to accurately infer the true labels $l$ through gradient inversion attacks (GI) \cite{2022Defending,kariyappa2023exploit} and residual reconstruction attacks (RR) \cite{tan2022residue}.

There is another type of attack that infers labels through the model training process. One possible method is that an adversary can obtain a small labeled dataset, supplement a top model to its pre-trained bottom model, and then proceed with training to achieve the goal of predicting labels. This type of attack is known as Passive Model Completion (PMC)  \cite{fu2022label}, where the passive party is semi-honest. An active version of model completion (AMC) is also proposed in \cite{fu2022label}, which utilizes malicious local optimizers instead of normal local optimizers. The success of Model Completion (MC) largely depends on the adequacy of the auxiliary data possessed by the attacke. Qiu et al. \cite{qiu2022your}, under the assumption that the attacker can access the global model and obtain prediction results, proposed a Label-related Relation Inference (LRI) attack targeting the label-related relationships within the graph owned by the active party.

However, these attacks are only applicable to a specific VFL scenario and a particular type of model, with some requiring auxiliary data as well. The LEA we propose is applicable to both SplitVFL and AggVFL and is effective on both models considered in this paper, which are logistic regression and neural network models. Some classic attacks and our proposed LEA are compared as shown in Table~\ref{tab:attacks}.

\section{Label Enumeration Attack}\label{sec:lea}
We propose a Label Enumeration Attack that enables passive parties to obtain private labels in VFL. Our key insight is that the local data of the passive party are inherently classifiable. LEA initiates by generating $n$ clusters through unsupervised clustering and copies $n!$ local models as simulated models. Then it enumerates $n!$ label permutations, assigning labels to each cluster in every simulated model for training. The attack model is identified by comparing the similarity with the normally trained model, which allows the adversary to independently utilize the attack model to complete the VFL process without the involvement of the active party's top model. We assess the model similarity between models on the cross-entropy loss function for both binary and multi-class tasks. The model with the highest match degree is selected as the attack model, which can independently and accurately recover the private labels of the label owner. The remainder of this section will provide a more detailed description of our proposed attack.

\begin{figure*}[htbp]
  \centering
  \includegraphics[width=\textwidth]{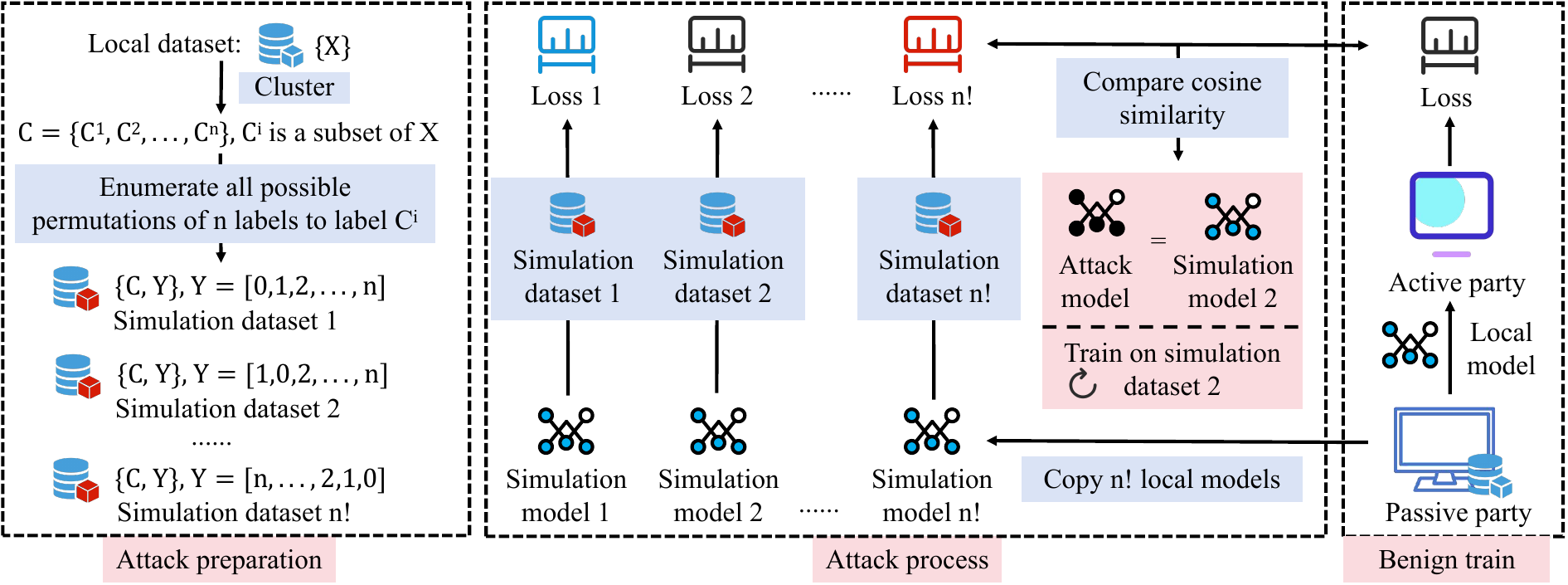}
  \caption{The attack process of LEA.}\label{attack}
\end{figure*}

Figure \ref{attack} illustrates the process of LEA. It is assumed that a passive party acts as an adversary hidden among all participating parties, aiming to steal the sample label information held by the active party. The adversary first prepares for the attack by clustering the local data features, grouping all samples into a set $C$. Each cluster $C^i$ in $C$ contains several samples, and then enumerating all possible $n!$ permutations of $n$ labels to assign labels to these clusters, thus equivalent to having $n!$ complete datasets. Next, the adversary duplicates $n!$ local models as simulated models (in the SplitVFL scenario, a top model is also simulated), and each trains on the $n!$ simulated datasets for one round to obtain $n!$ first-round loss gradients. At the same time, the adversary participates in normal training to obtain the normal first-round loss gradients. The adversary compares the loss values produced by each simulated model with the loss value generated by participating in the normal joint training using cosine similarity. The simulated model with the highest similarity in loss values is selected as the attack model, and is trained to convergence on its corresponding simulated dataset to predict the true labels. The following part of this chapter will provide a detailed introduction to the attack process.

\subsection{Threat Model}

We posits that each participant possesses a non-overlapping subset of features for the same set of samples, having completed the data alignment process inherent in VFL as in previous work \cite{fu2022label}. It is assumed that the features held by each participant are conducive to training a predictive model, implying that their local data are essentially classifiable. This assumption is justified, as participants can contribute to the global model only if they possess valuable features for VFL; otherwise, there is no merit in participating in the aggregation process. In scenarios with non-i.i.d. data, this paper considers the case where participants hold different proportions of sample features. We assume that the adversary has these abilities:

\begin{itemize}
    \item The adversary is aware of the type of training task and the number of labels. For example, with the CelebA dataset \cite{liu2018large}, the adversary knows that training tasks involve determining whether a person is wearing glasses or determining the gender, both of which have a label category count of 2. If it is about determining ethnicity, the category count is 3. This is often known in practice because, as the passive party in a VFL system, the adversary already needs to be informed about the training-related information.
    
    \item The adversary possesses the capability to perform clustering. This is the most critical assumption of this chapter, and subsequent experiments demonstrate that the effectiveness of the attack largely depends on the outcome of the clustering. Liu et al. \cite{lu2022dac} proposed a Deep Autoencoder-based Clustering (DAC) method that achieves an accuracy of 97.8\% on the MNIST dataset; Li et al. \cite{li2021contrastive} proposed a Contrastive Clustering (CC) method that achieves an accuracy of 79\% on the Cifar-10 dataset. Extant literature abounds with studies corroborating the feasibility of image clustering predicated upon feature-based categorization \cite{liu2022simplemkkm, wei2023learning, Seo2024, Ghoniem2024Improved, Li2024Ensemble, Nandi2024A, Chu2024Deep, Feifei2024Sub}. These clustering methods can all serve as components of the attack in this chapter, without being tied to any specific clustering method. 
\end{itemize}

It should be noted that, compared to previous label inference attacks, the adversary's assumption does not require an auxiliary dataset, nor does it require prior knowledge of the label distribution.

\subsection{Model Similarity}
In the VFL scenario, the three most commonly used models are logistic regression models, neural network models, and decision tree models \cite{liu2024vertical}. Our work focuses on the first two models. The logistic regression model is simple and efficient, used for dealing with binary classification problems, and is widely applied in fields such as financial risk assessment, credit scoring, disease diagnosis, and spam filtering. Its mathematical expression is:
\begin{equation}
	\sigma(z)=\frac{1}{1+e^{-z}}
\end{equation}
where $z$ is a linear combination of the input features, which can be expressed as $z=\beta_0+\beta_1x_1+\beta_2x_2+…+\beta_nx_n$, and $\beta_i$ are the weight parameters and $xi$ are the feature values. This model has a relatively small number of parameters and often converges to the unique minimum point of the loss function. Therefore, to compare the similarity between two logistic regression models, one must compare the similarity between their parameter vectors $\{\beta_i\}$. There are many metrics to determine the similarity of vectors, such as Euclidean distance, Manhattan distance, cosine similarity, Pearson correlation coefficient, KL divergence, and Chebyshev distance, etc. In the VFL scenario, the passive party can obtain the loss gradients updated by the model at each iteration, and thus knows the direction in which the model converges. For two models with the same initialization, if the direction of each update is similar, their final prediction results will also be similar. Therefore, we chooses cosine similarity, which is convenient for describing changes in direction, as the criterion for determining model similarity.

Neural network models have numerous local optima in their loss functions, and models with different initializations, even when trained on the same dataset, are likely to have dissimilar parameters. In SplitVFL, adversary cannot access the active party's top model, leading him/her to randomly generate a top model. This results in the simulated bottom model and the bottom model from joint training potentially converging to different local optima, manifesting as parameter dissimilarity. 

To address this, we found that comparing the cosine similarity of the first-round loss gradients rather than the parameters can assist the adversary in identifying the attack model corresponding to the true label sequence, when the sign of parameter initializations in both the active party's top model and the adversary's generated top model is the same (positive in this paper). We theoretically and empirically substantiate this approach in subsequent sections. As previously mentioned, SplitVFL is considered more secure than AggVFL, and correspondingly, the similarity in first-round loss gradients is more pronounced in AggVFL. We clearly demonstrate this in the experimental section.

\subsection{LEA for AggVFL}

In AggVFL, each passive party possesses a complete model. After the passive parties generate their output vectors, they transmit these to the active party. The active party then aggregates these vectors by summing them and inputs the combined vector into the loss function to calculate the loss with respect to the true label vector.

Before participating in global training, the adversary replicates $n!$ simulated models identical to the local model. Subsequently, the adversary performs clustering on the local data to obtain $n$ clusters. The adversary generates a set of labels, denoted as $A$, locally and then creates all possible permutations of this set, resulting in a collection of label sequences enumerated from the set $A$, which forms a set $\mathcal{L}$ of length $n!$ (where $n$ is the number of classes). We can confirm that there must be a sequence in $\mathcal{L}$ that corresponds to the real label of each cluster. Ordinarily, the passive party could not train models due to the lack of labels, but now, equipped with $n!$ label permutations, the adversary can assign labels to these $n$ clusters. Consequently, each simulated model is endowed with labeled training data. The model parameters, feature data, and training hyperparameters for these simulated models are identical, and the only variation is the labels that correspond to the data feature.

During the joint training process with the active party, each passive party sends their output results $\hat{y}_i=\theta_i(\mathcal{X}_i)$ to the active party, where $\theta$ denotes model parameters and $\mathcal{X} \in \mathbb{R}^{b \times f_i}$ denotes data features. The active party then sums and organizes these results and gets $\hat{y}=Sigmoid(\sum \hat{y}_i)$ where $\hat{y}$ consists of $m$ probabilities ($m$ means number of samples). The cross entropy loss function can be represented as:
\begin{equation}
    L=-\frac{1}{m}\sum_{i=1}^m\sum_{j=0}^n(y_{ij}\log \hat{y}_{ij})
\end{equation}
where $m$ denotes the number of samples, $y_{ij}$ is the probability that sample $i$ belongs to class $j$, with a value of 0 or 1. The first-round gradient of the output result for the adversary $\nabla \theta_a$ can be expressed as:
\begin{equation}\label{eq11}
\begin{aligned}
    \nabla \theta_a&= \frac{\partial L}{\partial \hat{y}} \frac{\partial \hat{y}}{\partial \hat{y}_a}\frac{\partial \hat{y}_a}{\partial \theta_a} \\
    &=-\frac{1}{m}\sum_{i=1}^m(y-\hat{y})\mathcal{X}_a
\end{aligned}
\end{equation}

For the adversary, multiple identical simulated models are trained individually. We assume that the sequence corresponding to the true labels in the set $\mathcal{L}$ is $l^*$, $l^*$ assigns the label vector $Y^*$ to each cluster, and each corresponding sample label is $y^*$. If the clustering accuracy is 100\%, then $y^*=y$, $\hat{y}'=Sigmoid(\theta_a(\mathcal{X}_a))=\hat{y}$. Assuming, as we do, that the dataset $\mathcal{X}_a$ is inherently separable and has been perfectly clustered with 100\% accuracy, then the corresponding labels $\hat{y}'$ and $y$ for any two samples within the same cluster would be equal. Therefore, $L'=L$ and the loss gradient for $y^*$ would be identical to the loss gradient from the joint training, resulting in the first-round loss gradients that is exactly the same as the one from the joint training:
\begin{equation}
\begin{aligned}
    \nabla \theta_a'&= \frac{\partial L'}{\partial \hat{y}'} \frac{\partial \hat{y}'}{\partial \theta_a} \\
    &=-\frac{1}{m}\sum_{i=1}^m(y^*-\hat{y}')\mathcal{X}_a \\
    &=\nabla \theta_a
\end{aligned}
\end{equation}

Although current clustering algorithms do not achieve 100\% accuracy, they can reach satisfactory precision levels. Therefore, while there may be some discrepancy in the loss gradients, the loss gradient for $y^*$ will be the smallest compared to other label sequences. Consequently, the simulated model trained with $y^*$ will be the most similar to the model produced by the joint training process and can be used to attack for predicting labels.

Furthermore, comparing similarities using the first-round loss gradients rather than the final model has two advantages: (1) It significantly reduces computational costs. Identifying the attack model with high probability after the first round of iteration allows for the selection of the most similar or top-ranking models for further training, while discarding the subsequent training of other simulated models, resulting in a reduction of computational costs by more than 100 times when the value of $n$ is greater than 5. (2) Clear differences in similarity. After the first round of iteration, the differences between the simulated models and the federated training model are pronounced, except for the attack model. If training is continued until convergence, the differences in parameters among numerous simulated models may actually decrease, which is not conducive to making judgments.

As discussed earlier, the passive party in the VFL process can obtain the loss gradients of the model after each update, knowing the direction of model convergence. Therefore, for two models with the same initialization, if the direction of each update is similar, the final models obtained will also be similar. Consequently, cosine similarity, which is convenient for describing changes in direction, is chosen as the metric for determining model similarity:
\begin{equation}
	Cosine\_Similarity(\nabla \theta_a',\nabla \theta_a)=\frac{\nabla \theta_a'\cdot\nabla \theta_a}{\|\nabla \theta_a'\|\|\nabla \theta_a\|}
\end{equation}

The closer the result is to 1, the higher the similarity between the two variables, and vice versa, the closer it is to -1, the lower it is.

It should be noted that whether the adversary is constructing a complete dataset, simulating label sequences, generating and training simulated models, or determining similarity to identify the attack model and using the attack model to predict sample labels, all these processes are carried out locally without any interference in the global training process. Therefore, for the active party, they cannot detect any attack, let alone detect and exclude it. This meets the high concealment requirements of attacks, allowing adversaries to steal label privacy information without being noticed.

\subsection{LEA for SplitVFL}
In SplitVFL, both the passive parties and active party possess portions of the global model, and it is only through their collaboration that the complete forward and backward propagation processes can be fulfilled. This approach differs from AggVFL in a significant way: upon receiving the output results from the passive parties, the active party concatenates these results rather than summing them. This concatenated output is then fed into the top model to obtain the predictive values. As a result of this concatenation, the number of parameters in the top model is proportional to the number of passive parties involved. Each passive party contributes a unique part of the overall feature space, and the top model integrates these parts to perform the final prediction.

In the context with $K$ passive parties, the active party's top model can be denoted by a simple expression to illustrate the process:
\begin{equation}
    \theta_T=(\theta_T^1,\theta_T^2,...,\theta_T^K)
\end{equation}
where $\theta_T$ denotes top model. When the active party receives output results from the bottom models of the passive parties, he/she concatenates these results and gets $\mathcal{Y}$, then feeds into the top model. The output $P$ of the top model, represents the probability of a sample corresponding to each label. This process is akin to computing the sum:
\begin{equation}
\begin{aligned}
    P&=Sigmoid(\theta_T(\mathcal{Y})) \\
    &=Sigmoid(\theta_T^1(\hat{y}_1)+\theta_T^2(\hat{y}_2)+...+\theta_T^K(\hat{y}_K))
\end{aligned}
\end{equation}
which signifies the aggregation of the features from all passive parties. The subsequent calculation of the loss gradient will differ slightly from that in AggVFL:
\begin{equation}\label{eq15}
\begin{aligned}
    \nabla \theta_{ba}&= \frac{\partial L}{\partial P} \frac{\partial P}{\partial \hat{y}} \frac{\partial \hat{y}}{\partial \hat{y}_a}\frac{\partial \hat{y}_a}{\partial \theta_{ba}}\\
    &=-\frac{1}{m}\sum_{i=1}^m(y-\hat{y})\theta_T^a\mathcal{X}_a
\end{aligned}
\end{equation}
where $\theta_{ba}$ denotes the the the bottom model. In the scenario described, the adversary, which is a passive party, only has the bottom model $\theta_a$. Even with an enumerated set of labels $\mathcal{L}$, the adversary cannot complete the training on its own. Consequently, the adversary adds an output layer $\theta_t$ to complement the bottom model, thus forming a complete trainable model $\theta=(\theta_a,\theta_t)$. We designate $\theta_t$ as a fully connected layer, with the input dimension matching the characteristic dimension of the output of $\theta_a$, and the output dimension equal to the number of labels $n$. In this context, besides generating the same $A$, $\mathcal{L}$, $C$ as in AggVFL, the adversary also has to create the top model $\theta_t$ to concatenate with $\theta_a$. Furthermore, the adversary replicates $n!$ complete simulated models, each with $\theta_t$ added to $\theta_a$, to carry out the training process. For the adversary, $\hat{y}_a=Sigmoid(\theta_t(\theta_{ba}(\mathcal{X}_a)))$, the loss gradient corresponding to the model that matches the true label sequence $l^*$ can be represented as:
\begin{equation}
\begin{aligned}
    \nabla \theta_{ba}&= \frac{\partial L'}{\partial P'} \frac{\partial P'}{\partial \hat{y}_a} \frac{\partial \hat{y}_a}{\partial \theta_{ba}}\\
    &=-\frac{1}{m}\sum_{i=1}^m(l^*-\hat{y}_a)\theta_t\mathcal{X}_a
\end{aligned}
\end{equation}
where $L'$ and $P'$ are the loss function and intermediate output corresponding to $l^*$, respectively. Indeed, while the adversary does not have access to the parameters $\theta_T^a$ of the active party's top model, if the top model's parameters are initialized to be greater than zero, the adversary can initiate the top layer of the simulated models with positive values as well to carry out the attack. In the case of a Logistic Regression model, the parameters of the top model correspond to the number of passive parties, and for a specific passive party, $\theta_T^a$ is a definite value. The attacker employs $n!$ simulated models, each augmented with a top layer $\theta_t$, and trains them under different label permutations. Even if there are differences in $\theta_T^a$ and $\theta_t$, as long as the signs are the same, the model corresponding to $l^*$ (the true label sequence) will have a bottom model that is most similar to the joint training bottom model in terms of parameters. In the context of Neural Network models, the parameter count of a single layer of the top model is the product of the number of passive parties and the feature dimension of the intermediate results. For a specific passive party, $\theta_T^a$ is a one-dimensional vector. It is easy for the sign characteristics to be lost in subsequent computations, hence we require that the parameters are initialized to be positive. After the model corresponding to $l^*$ has converged, it remains the most similar to the joint training model in terms of parameters. We provide a detailed process in Algorithm ~\ref{alg:AggLEA}.

\begin{algorithm}[htbp]
\centering
\caption{Label Enumeration Attack in VFL.}\label{alg:AggLEA}
\begin{algorithmic}
        \renewcommand{\algorithmicrequire}{\textbf{Input:}}
        \renewcommand{\algorithmicensure}{\textbf{Output:}}
        \REQUIRE Number of labels $n$, adversary's local model $\theta_a$, adversary's first-round loss gradients $\nabla \theta_a$ in AggVFL or $\nabla \theta_{ba}$ in SplitVFL, adversary's local data $\mathcal{X}_a \in \mathbb{R}^{b \times f_a}$, training epochs $e$, learning rate $\eta$.
        \ENSURE Attack model $\theta_I$.
    \end{algorithmic}
    \begin{algorithmic}
        \STATE $A \leftarrow (0,1,2,...,n-1)$
        \STATE $\mathcal{L} \leftarrow$ Generate $n!$ permutations of set $A$.
        
        \STATE $\{\mathcal{C}^i\}_{i=1}^n \leftarrow Clustering(\mathcal{X}_a)$
        \STATE $\{Simulate\_label_i\}_{i=0}^{n!-1} \leftarrow$ Label the cluster $\mathcal{C}^x$ as $\mathcal{L}_i^x$.
        \STATE $\{\theta_i\}_{i=0}^{n!-1} \leftarrow$ Copy $n!$ models that are the same as local model $\theta_a$ \, //AggVFL
        \STATE $\theta_t \leftarrow$ Generate a top model.\quad //SplitVFL
        \STATE $\{\theta_i\}_{i=0}^{n!-1} \leftarrow$ Copy $n!$ bottom models that are the same as local model $\theta_a$ and complete the model with $\theta_t$.\quad //SplitVFL
        \FOR {$epoch \in \{0,1,...,e\}$}
        \FOR {$i \in$ index of $\{\theta_i\}_{i=0}^{n!-1}$}
        \STATE $\hat{y}_i \leftarrow Forward(\theta_i,\mathcal{C})$
        \STATE $L \leftarrow LossFunction(Sigmoid(\hat{y}_i),Simulate\_label_i)$
        \STATE $\nabla \theta_i \leftarrow \frac{\partial L}{\partial \hat{y}_i}\frac{\partial \hat{y}_i}{\partial \theta_i}$\quad //AggVFL
        \STATE $\nabla \theta_{bi} \leftarrow \frac{\partial L}{\partial \hat{y}_i}\frac{\partial \hat{y}_i}{\partial \theta_t}$\quad //SplitVFL
        \STATE $\theta_i \leftarrow \theta_i-\eta \times \nabla \theta_i$
        \IF{$epoch =0$}
        \STATE $Score_i \leftarrow Cosine\_Similarity(\nabla \theta_i,\nabla \theta_a)$\quad //AggVFL
        \STATE $Score_i \leftarrow Cosine\_Similarity(\nabla \theta_{ba}, \nabla \theta_{bi})$\, //SplitVFL
        \ENDIF
        \ENDFOR
        \IF{$epoch =0$}
		\STATE $I \leftarrow$ Index of $Max(\{Score_i\}_{i=0}^{n!-1})$
		\STATE $\{\theta_i\}_{i=0}^{n!-1}\leftarrow \{\theta_I\}$
		\ENDIF
        \ENDFOR
        \RETURN Attack model $\theta_I$.
    \end{algorithmic}
\end{algorithm}

\subsection{Binary-LEA}
Given that training $n!$ models can be computationally intensive, especially when $n$ is large (for instance, $n=10$ results in $n!=3,628,800$), we propose a Binary Label Enumeration Attack (Binary-LEA) to reduce the computational overhead.


We transform a multi-class classification task into $\lfloor \frac{n}{2} \rfloor$ binary classification tasks, thereby generating $\lfloor \frac{n}{2} \rfloor$ attack models. From the clusters formed by clustering, we select two clusters at a time, which results in $N(N-1)$ possible label permutations for labels $l_1$ and $l_2$ (where $N$ is the number of remaining labels after accounting for the two selected clusters). Using the previously described method, we derive one attack model from the $N(N-1)$ simulated models. The output of this model includes three values: the probability of label $l_1$, the probability of label $l_2$, and the probability that the sample does not belong to either $l_1$ or $l_2$. With each generated attack model, the value of $N$ is decremented by 2. We provide a detailed process in Algorithm ~\ref{alg:Binary-LEA}. So the total number of permutations decreased from $n!$ to $n(n-1)+(n-2)(n-3)+...+2\times1$ which is equivalent to the order of magnitude of $\mathcal{O}(n^3)$. Further simplification can lead to:

\begin{equation}
    n(n-1)+(n-2)(n-3)+...+2\times1\\= \frac{n(n+2)(2n-1)}{12}
\end{equation}

When making predictions, the attacker inputs data into these $\lfloor \frac{n}{2} \rfloor$ attack models and determines the label value by synthesizing their output results. This approach reduces the number of models that need to be trained from $\mathcal{O}(n!)$ to $\mathcal{O}(n^3)$, which significantly decreases the computational burden when $n>3$.

\begin{algorithm}[htbp]
\centering
\caption{Binary-Label Enumeration Attack (Binary-LEA).}\label{alg:Binary-LEA}
\begin{algorithmic}
        \renewcommand{\algorithmicrequire}{\textbf{Input:}}
        \renewcommand{\algorithmicensure}{\textbf{Output:}}
        \REQUIRE Number of labels $n$, adversary's local model $\theta_a$, adversary's first-round loss gradients $\nabla \theta_a$, adversary's local data $\mathcal{X}_a \in \mathbb{R}^{b \times f_a}$, training epochs $e$, learning rate $\eta$.
        \ENSURE Attack models $\theta_1,\theta_2,...,\theta_{\lfloor \frac{n}{2} \rfloor}$.
    \end{algorithmic}
    \begin{algorithmic}
        
        \STATE $N \leftarrow n$ 
        \STATE $A \leftarrow (0,1,2,...,n-1)$
        \STATE $\{\mathcal{C}^i\}_{i=1}^n \leftarrow Clustering(\mathcal{X}_a)$
        \FOR {$l \in \{0,1,...,\lfloor \frac{n}{2} \rfloor\}$}
        \STATE $\mathcal{C}^1,\mathcal{C}^2 \leftarrow$ Select two clusters of $\{\mathcal{C}^i\}_{i=1}^n$ in order.
        \STATE $\mathcal{L} \leftarrow$ Generate $N(N-1)$ permutations of random two labels $l^1,l^2$ of set $A$.
        \STATE $\{\theta_i\}_{i=0}^{N(N-1)-1} \leftarrow$ Copy $N(N-1)$ models that are the same as local model $\theta_a$\quad //AggVFL
        \STATE $\theta_t \leftarrow$ Generate a top model.\quad //SplitVFL
        \STATE $\{\theta_i\}_{i=0}^{n!-1} \leftarrow$ Copy $N(N-1)$ bottom models that are the same as local model $\theta_a$ and complete the model with $\theta_t$.\quad //SplitVFL
        
        \STATE $Simulate\_label \leftarrow$ Label the cluster $\mathcal{C}^x$ with $l^1,l^2$ as $\mathcal{L}_i^x$.
        \FOR {$epoch \in \{0,1,...,e\}$}
        \FOR {$i \in$ index of $\{\theta_i\}_{i=0}^{N(N-1)-1}$}
        
        \STATE $\hat{y}_i \leftarrow Forward(\theta_i,\mathcal{C})$
        \STATE $L \leftarrow LossFunction(Sigmoid(\hat{y}_i),Simulate\_label_i)$
        \STATE $\nabla \theta_i \leftarrow \frac{\partial L}{\partial \hat{y}_i}\frac{\partial \hat{y}_i}{\partial \theta_i}$\quad //AggVFL
        \STATE $\nabla \theta_{bi} \leftarrow \frac{\partial L}{\partial \hat{y}_i}\frac{\partial \hat{y}_i}{\partial \theta_t}$\quad //SplitVFL
        \STATE $\theta_i \leftarrow \theta_i-\eta \times \nabla \theta_i$ 
        \IF{$epoch=0$}
        \STATE $Score_i \leftarrow Cosine\_Similarity(\nabla \theta_i,\nabla \theta_a)$\quad //AggVFL
        \STATE $Score_i \leftarrow Cosine\_Similarity(\nabla \theta_{ba}, \nabla \theta_{bi})$\, //SplitVFL
        \ENDIF
        \ENDFOR
        \IF{$epoch =0$}
		\STATE $I \leftarrow$ Index of $Max(\{Score_i\}_{i=0}^{N(N-1)-1})$
		\STATE $\{\theta_i\}_{i=0}^{N(N-1)-1}\leftarrow \{\theta_I\}$
        \ENDIF
        \ENDFOR
        \STATE $\theta_l \leftarrow$ Retrain $\theta_I$ to additionally output the probability that neither of these two categories is true.
        \STATE $N \leftarrow N-2$
        \STATE $\{\mathcal{C}\} \leftarrow \{\mathcal{C}\}-\mathcal{C}^1-\mathcal{C}^2$
        \ENDFOR
        \RETURN Attack models $\theta_1,\theta_2,...,\theta_{\lfloor \frac{n}{2} \rfloor}$.
    \end{algorithmic}

\end{algorithm}

\section{Experiments}
The experiments were conducted on a Microsoft Windows 11 Pro operating system, version 22635.2483, with a host processor of 13th Gen Intel(R) Core(TM) i7-13700F, 2100MHz, featuring 16 cores and 24 logical processors.

\subsection{Experimental Setup}
\subsubsection*{Datasets}
The Breast Cancer dataset \cite{Amala2022Breast} comprises 569 samples with 30 features. The target variable is binary, indicating whether the tumor is benign or malignant. The Give-me-some-credit dataset \cite{Bradford2023} has 150,000 samples with a feature size of 10, used to predict whether individuals meet the loan requirements of financial institutions. To explore the effectiveness of LEA on multiclass classification tasks, we choose to conduct experiments on the MNIST dataset \cite{Wang2020Improvement}. The MNIST training dataset contains 60,000 samples, each sample being a 28x28 pixel grayscale image of a handwritten digit from 0 to 9. We select 3, 5, and 10 digits for multiclass classification tasks to observe the attack effects. 

\subsubsection*{Models}
On the Breast Cancer and the Give-me-some-credit dataset, we employed both a Logistic Regression model and a MLP model with three hidden layers. For the MNIST dataset, we utilized the residual network ResNet18 model \cite{Jian2016Deep}. In the context of SplitVFL, the number of parameters for the top model corresponded to the count of passive parties involved. In the SplitVFL scenario, the bottom model encompassed the parameters of the first two layers of the MLP model, while the third layer, which operates as a standalone entity, served as the top model. The ResNet18 model is divided into a bottom model from the input layer to the third residual block, and the remaining part is the top model.

\subsubsection*{VFL settings}
To simulate VFL scenarios, we partition the datasets as follows. In Breast Cancer dataset, which has 30 features, we have configured settings for two-party, three-party, and four-party scenarios, with the distribution of feature numbers being 28-2, 19-9-2, and 6-13-9-2 respectively (the attacker possesses the first number of features, and the active party holds the last number of features). For the Give-me-some-credit dataset, we divide the features into 5-5. These divisions corresponds to the real-world scenario where the passive party often possesses more feature information than the active party, while also considering the situation where there are fewer but more important features. For MNIST dataset, we consider a two-party scenario where each image is split into two halves, distributed among the two participating parties, a setup similar to that in work \cite{fu2022label}. From the dataset, we have selected samples labeled 0, 1, and 2 for a ternary classification task named MNIST-3, samples labeled 0, 1, 2, 3, and 4 for a quinary classification task named MNIST-5, and samples containing all labels for a decaclassification task named MNIST-10.

\subsection{Attack Evaluation}
\begin{figure*}[htbp]
  \centering
  \includegraphics[width=\textwidth]{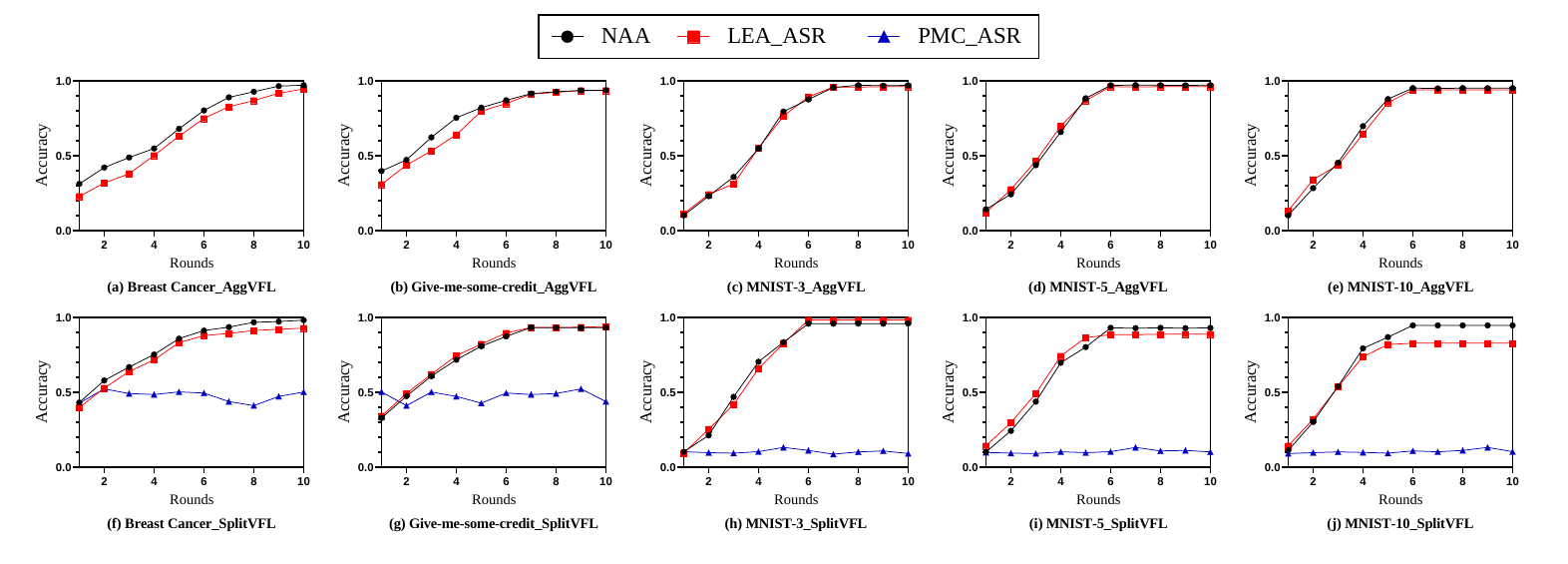}
  \caption{Comparison of attack effects of LEA and PMC using NN models in SplitVFL and AggVFL scenarios under five dataset settings.}\label{LEAPMC}
\end{figure*}
As stated above, we conducted our attack five times under each VFL setting, and the average results are presented in Table~\ref{tab：attack performance}. We measure the No Attack Accuracy (NAA), Cluster Accuracy (CA) and Attack Success Rate (ASR) of LEA over the testing data. We define ASR as the Top-1 accuracy of the attack model on the validation dataset. We expand the first-round loss gradients into a one-dimensional vector and then calculate the cosine similarity. Our attacks achieved good performance on both the Breast Cancer and Give-me-some-credit datasets, with attack accuracies exceeding 0.9. For the multi-class task on MNIST, our attacks also performed well, maintaining an accuracy above 0.8. It can be observed that the cluster accuracy directly influences the attack accuracy, which aligns with our expectations that the performance of the attack model is inextricably linked to the accuracy of the clusters formed by the clustering process.

\begin{table*}[htbp]
\renewcommand{\arraystretch}{1}
\caption{The attack performance of LEA.}
\begin{center}
\addtolength{\tabcolsep}{0pt}
\begin{tabular}{c|c|c|c|c|c|ccc}
\hline
\multirow{3}{*}{\textbf{Dataset}} &\multirow{3}{*}{\textbf{Scene}}& \textbf{Number} & \textbf{Number of} & \textbf{Distribution} &\multirow{3}{*}{\textbf{Model}} & \multicolumn{3}{c}{\multirow{2}{*}{\textbf{Attack Performance}}} \\ 
& &\textbf{of} &\textbf{Simulated} &\textbf{of}& & & \\
\cline{7-9} 
& & \textbf{Classes} &\textbf{ Models}  & \textbf{Features} & &\textbf{\textbf{NAA}}& \textbf{\textbf{CA}}& \textbf{\textbf{ASR$^{\mathrm{1}}$}} \\
 
\hline
\multirow{12}{*}{Breast Cancer}  &\multirow{6}{*}{AggVFL}& \multirow{6}{*}{2} &\multirow{6}{*}{2} & \multirow{2}{*}{\textbf{28}-2} &LR & 0.970 & 0.932 & 0.946\\
& & & & & MLP & 0.982 & 0.932 & 0.938 \\
\cline{5-9} 
& & & & \multirow{2}{*}{\textbf{19}-9-2} &LR & 0.963 & 0.921 & 0.965 \\
& & & & & MLP & 0.973 & 0.921 & 0.947 \\
\cline{5-9} 
& & & & \multirow{2}{*}{\textbf{6}-13-9-2} &LR & 0.912 & 0.932 & 0.947 \\
& & & & & MLP & 0.973 & 0.932 & 0.921 \\
\cline{2-9} 
  &\multirow{6}{*}{SplitVFL}& \multirow{6}{*}{2} &\multirow{6}{*}{2} & \multirow{2}{*}{\textbf{28}-2} &LR & 0.947 & 0.932 & 0.956\\
& & & & & MLP & 0.982 & 0.932 & 0.929 \\
\cline{5-9} 
& & & & \multirow{2}{*}{\textbf{19}-9-2} &LR & 0.945 & 0.921 & 0.964 \\
    & & & & & MLP & 0.981 & 0.921 & 0.956 \\
\cline{5-9} 
& & & & \multirow{2}{*}{\textbf{6}-13-9-2} &LR & 0.922 & 0.932 & 0.947 \\
& & & & & MLP & 0.968 & 0.932 & 0.915 \\
\hline
\multirow{4}{*}{Give-me-some-credit}& \multirow{2}{*}{AggVFL} & \multirow{2}{*}{2} & \multirow{2}{*}{2} & \multirow{2}{*}{\textbf{5}-5} &LR & 0.837 & 0.935 & 0.839 \\
& &	& & & MLP & 0.936 & 0.935 & 0.936 \\
\cline{2-9} 
& \multirow{2}{*}{SplitVFL} & \multirow{2}{*}{2} & \multirow{2}{*}{2} & \multirow{2}{*}{\textbf{5}-5} &LR & 0.842 & 0.935 & 0.812 \\
& &	& & & MLP & 0.933 & 0.935 & 0.940 \\
\hline
\multirow{2}{*}{MNIST-3}& AggVFL& \multirow{2}{*}{3} &\multirow{2}{*}{6} & \multirow{2}{*}{\textbf{18x28}-10x28} &\multirow{2}{*}{ResNet18}& 0.970 & 0.930 & 0.963 \\
& SplitVFL&  &  & & & 0.960 & 0.930 & 0.983 \\
\hline
MNIST-5& AggVFL&\multirow{2}{*}{5} & \multirow{2}{*}{26} &  \multirow{2}{*}{\textbf{18x28}-10x28} &\multirow{2}{*}{ResNet18}& 0.972 & 0.912 & 0.963 \\
(Adopt Binary-LEA)& SplitVFL&  &  &  & & 0.930 & 0.912 & 0.889 \\
\hline
MNIST-10 & AggVFL&  \multirow{2}{*}{10} & \multirow{2}{*}{190} & \multirow{2}{*}{\textbf{18x28}-10x28} &\multirow{2}{*}{ResNet18}& 0.952 & 0.887 & 0.940 \\
(Adopt Binary-LEA)& SplitVFL&  &  &  & & 0.947 & 0.887 & 0.828 \\
\hline
\multicolumn{6}{l}{$^{\mathrm{1}}$We take the Top-1 accuracy of the attack model in the validation set as ASR.}
\end{tabular}
\label{tab：attack performance}
\end{center}
\end{table*}

\begin{figure}[htbp]
  \centering
  \includegraphics[width=\linewidth]{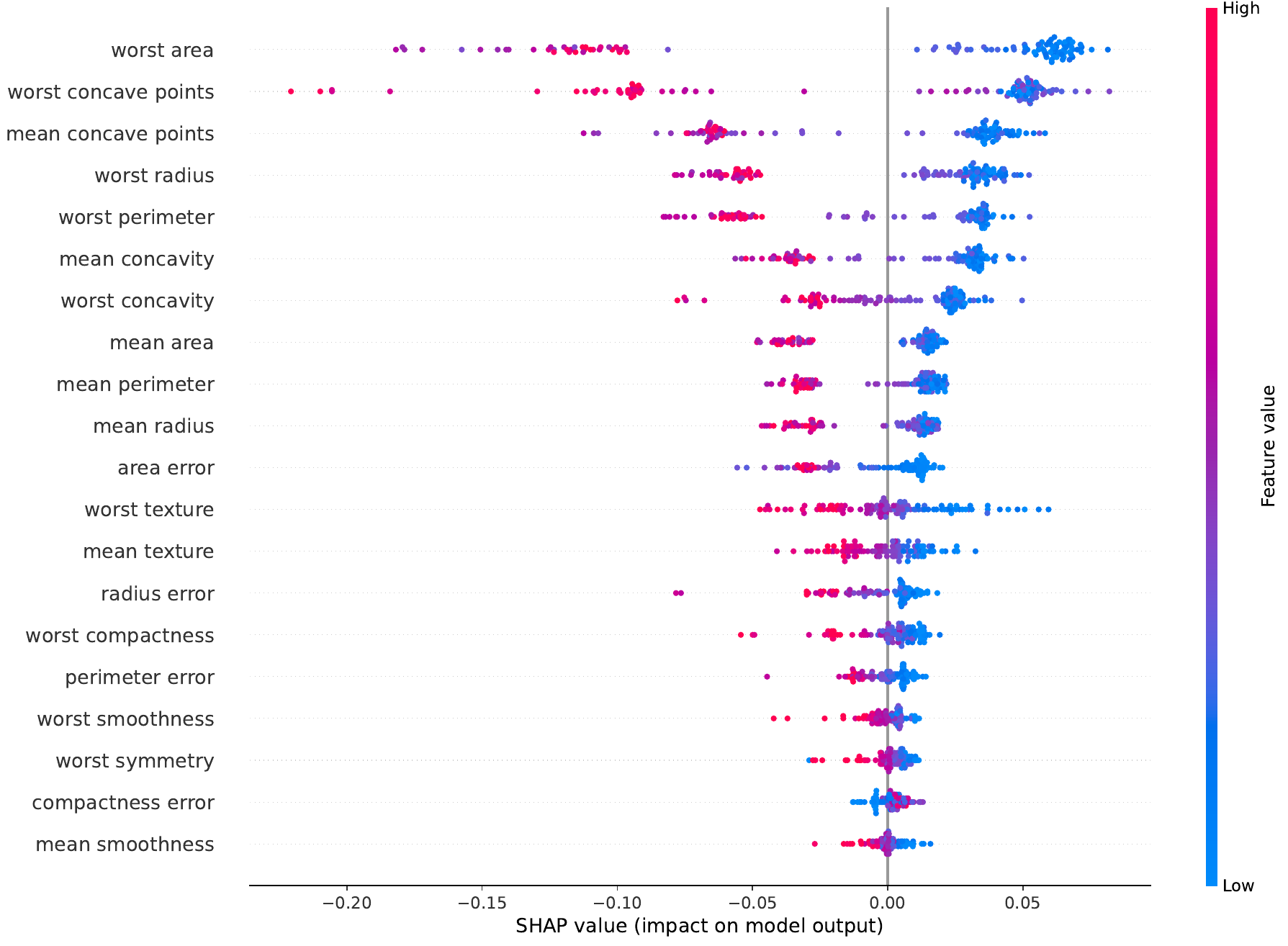}
  \caption{The SHAPLEY Value Analysis Chart for the Breast Cancer Dataset.}\label{shap_bc}
\end{figure}

\begin{figure}[htbp]
  \centering
  \includegraphics[width=\linewidth]{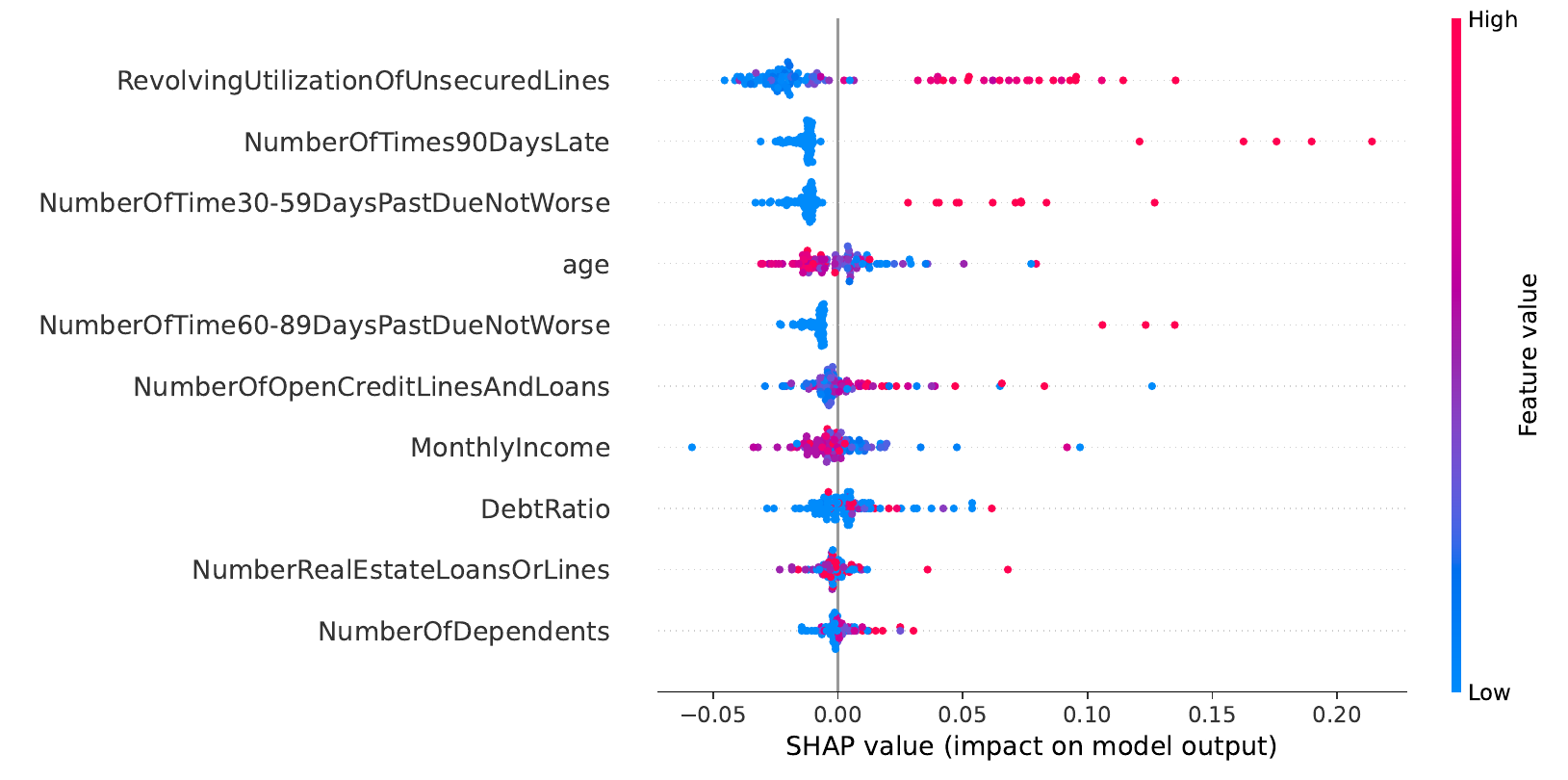}
  \caption{The SHAPLEY Value Analysis Chart for the Give-me-some-credit Dataset.}\label{shap_gmsc}
\end{figure}

We conducted SHAPLEY value analysis on the Breast Cancer and Give-me-some-credit datasets, as shown in Figure~\ref{shap_bc} and Figure~\ref{shap_gmsc}. The top-ranked features in these two datasets play a significant role in the prediction results, which confirms that in VFL, the passive party can achieve clustering with a small number of useful features.

We compared the attack effectiveness of LEA with one of the state-of-the-art label inference attacks, passive model completion (PMC) \cite{fu2022label}, on five datasets using neural network models in both splitVFL and AggVFL scenarios. However, due to the heavy reliance of PMC's attack effectiveness on auxiliary datasets not considered in this paper, it exhibited a form of random guessing, and because it is only applicable to SplitVFL, we did not compare it in AggVFL. From Figure \ref{LEAPMC}, it can be seen that the attack accuracy of LEA approaches the test accuracy of models trained normally without attacks in both five datasets and two VFL algorithm scenarios.

We conducted a comparative analysis of the cosine similarity between the model parameters and the first-round loss gradients in the binary classification tasks for both the simulated and the joint training model. As can be observed from Table~\ref{tab：cosine similarity}, in SplitVFL, comparing the cosine similarity of the parameters can potentially lead to the selection of an incorrect attack model, whereas comparing the similarity of the first-round loss gradients effectively identifies the correct attack model. This aligns with the analysis presented in Section~\ref{sec:lea}. For AggVFL, when using the NN model, the similarity in parameters is not distinctly indicative, whereas the first-round loss gradients provide a clear determination of which simulated model corresponds to the attack model. This confirms the feasibility of using the cosine similarity of the first-round loss gradients between models to ascertain model similarity.

\begin{table}[htbp]
\renewcommand{\arraystretch}{1.1}
\setlength{\tabcolsep}{3pt}
\caption{Comparison of cosine similarity between parameters and first-round loss gradients on Breast Cancer dataset.}
\begin{center}

\begin{tabular}{c|c|c|c|c|c}
\hline 
\multirow{3}{*}{\textbf{Scenario}} & \multirow{3}{*}{\textbf{Model}} & \multicolumn{4}{c}{\textbf{Cosine Similarity}} \\
\cline{3-6}
& & \multicolumn{2}{c|}{\textbf{Parameters}}&\multicolumn{2}{c}{\textbf{First-round loss gradients}}\\
\cline{3-6}
& & \textbf{Attack}&\textbf{Another$^{\mathrm{1}}$}& \textbf{Attack}&\textbf{Another}\\
\hline
\multirow{2}{*}{AggVFL} & LR& 0.908  & -0.161 & 0.996  & -0.438 \\
& NN& 0.945 & 0.840 & 0.982  & -0.967 \\
\multirow{2}{*}{SplitVFL} & LR& 0.705  & 0.946 & 0.972  & -0.723  \\
& NN& 0.746 & 0.690 & 0.871  & -0.695  \\
\hline
\multicolumn{6}{l}{$^{\mathrm{1}}$Another is the model corresponding to the wrong label sequence.}
\end{tabular}
\label{tab：cosine similarity}
\end{center}
\end{table}

\begin{figure}[htbp]
  \centering
  \includegraphics[width=\linewidth]{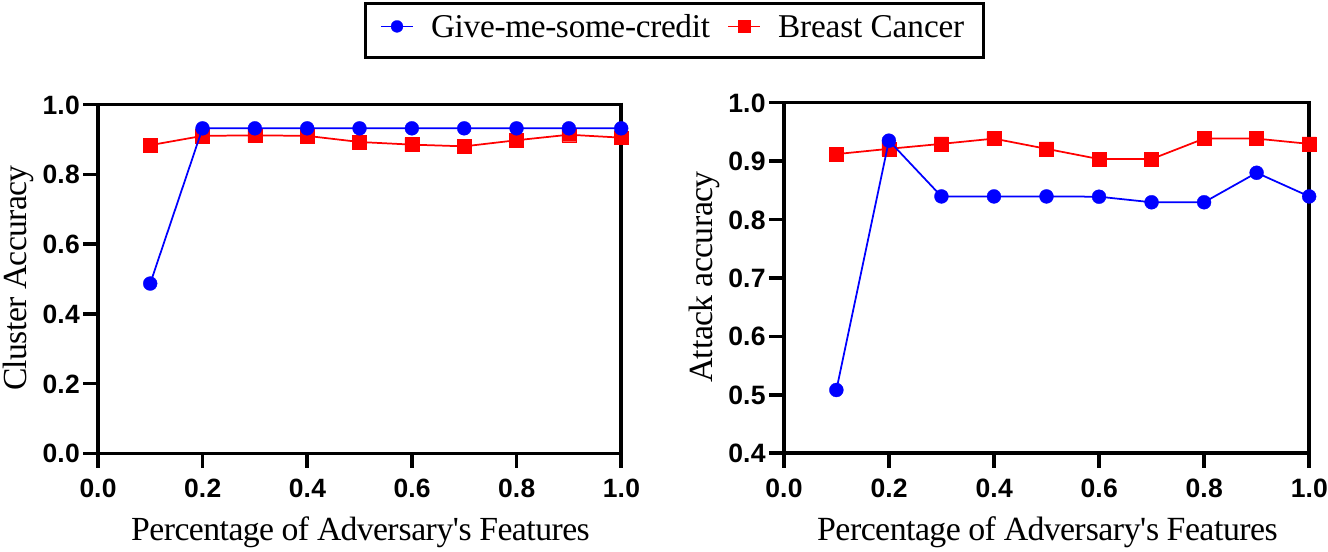}
  \caption{The impact of different percentage of adversary's features on cluster accuracy and attack performance in Binary SplitVFL with NN model.}\label{Cluster}
\end{figure}

\textbf{Sensitivity to Cluster Accuracy: }The experimental results indicate a strong correlation between ASR and CA. As shown in Figure \ref{Cluster}, the experiment simulated scenarios where the adversary has different proportions of features for two binary classification tasks, the Breast Cancer dataset and the Give-me-some-credit dataset. For the Breast Cancer dataset, even when the adversary holds only 10\% of the features, that is, 3 feature values, the clustering accuracy can still reach 88.4\%, enabling the attack model to achieve a high prediction accuracy of 91.2\%. In contrast, for the Give-me-some-credit dataset, when the adversary holds 10\% of the features, i.e., 1 feature value, the clustering accuracy is 48.7\%, resulting in an attack accuracy of only 50.8\%. This is because for this dataset, a single feature does not sufficiently classify the samples. When holding more than two feature values, the clustering accuracy stabilizes at 93.2\%, and the attack accuracy also increases to around 85.7\%.

These results demonstrate that the proportion of features the adversary possesses is not the most critical factor. Instead, the quality of the adversary's data and its ability to support high cluster accuracy are the key determinants of the success of LEA. The higher the cluster accuracy, the more effective the attack, as it can more closely approximate the true label distribution and consequently predict the labels of the samples with greater accuracy.

\begin{table}[htbp]
\renewcommand{\arraystretch}{1}
\setlength{\tabcolsep}{2pt}
\caption{The time cost of Binary-LEA and LEA under various dataset settings in splitVFL scenario.}
\begin{center}
\begin{tabular}{ccc}
\hline 
\textbf{Dataset}& \textbf{Scheme} & \textbf{Time cost (s)} \\
\hline
Breast Cancer& LEA & 0.3960\\
Breast Cancer& Binary-LEA & 0.4010\\
Give-me-some-credit& LEA & 12.3576\\
Give-me-some-credit& Binary-LEA & 13.6124\\
MNIST-3 & LEA& 15.1794 \\
MNIST-3 & Binary-LEA& 16.1245 \\
MNIST-5 & LEA& 1254.2317 \\
MNIST-5 & Binary-LEA& 256.4912 \\
MNIST-10 & LEA& - \\
MNIST-10 & Binary-LEA& 4924.3223 \\
\hline
\end{tabular}
\label{timecost}
\end{center}
\end{table}

We tested the time cost of Binary-LEA and LEA under various dataset settings in the splitVFL scenario, as shown in table \ref{timecost}. When the number of dataset labels is less than or equal to 3, these two schemes generate the same number of simulated models, so their time overheads are very similar. However, due to the different sizes of the datasets, the attack time costs vary between datasets. When the number of dataset labels exceeds 3, as can be seen from MNIST-5, Binary-LEA significantly reduces the number of simulated models generated, greatly reducing the time cost compared to LEA. For MNIST-10, the number of simulated models generated by LEA is theoretically 19,098 times that of Binary-LEA, so the time overhead is approximately 19,098 times that of Binary-LEA. Since Binary-LEA takes 4,924.3223 seconds, the time required for LEA would be about three years, hence this paper does not test it here.

LEA covers a broader range of VFL scenarios and requires less auxiliary data, while still achieving effective label inference attack results. This is because, unlike previous work, the effectiveness of LEA is primarily dependent on the adversary's ability to cluster local data effectively.

\section{Defenses}
LEA leverages information from the loss gradients provided by the active party. Thus, defensive methods applied to the gradients might mitigate our attack. We consider two methods: (\textit{i}) Noisy Gradients \cite{zhu2019deep}: Adding Laplace noise to the loss gradients before transmission to protect gradient privacy. (\textit{ii}) Gradient Compression \cite{kairouz2021advances}: Sending only a proportion of the most significant gradients (those with larger absolute values), which can still yield a highly accurate model. These methods can cause the adversary to select an incorrect attack model by altering the gradient information. Additionally, we propose a defense mechanism based on a label mapping table.

\begin{figure}[htbp]
  \centering
  \includegraphics[width=\linewidth]{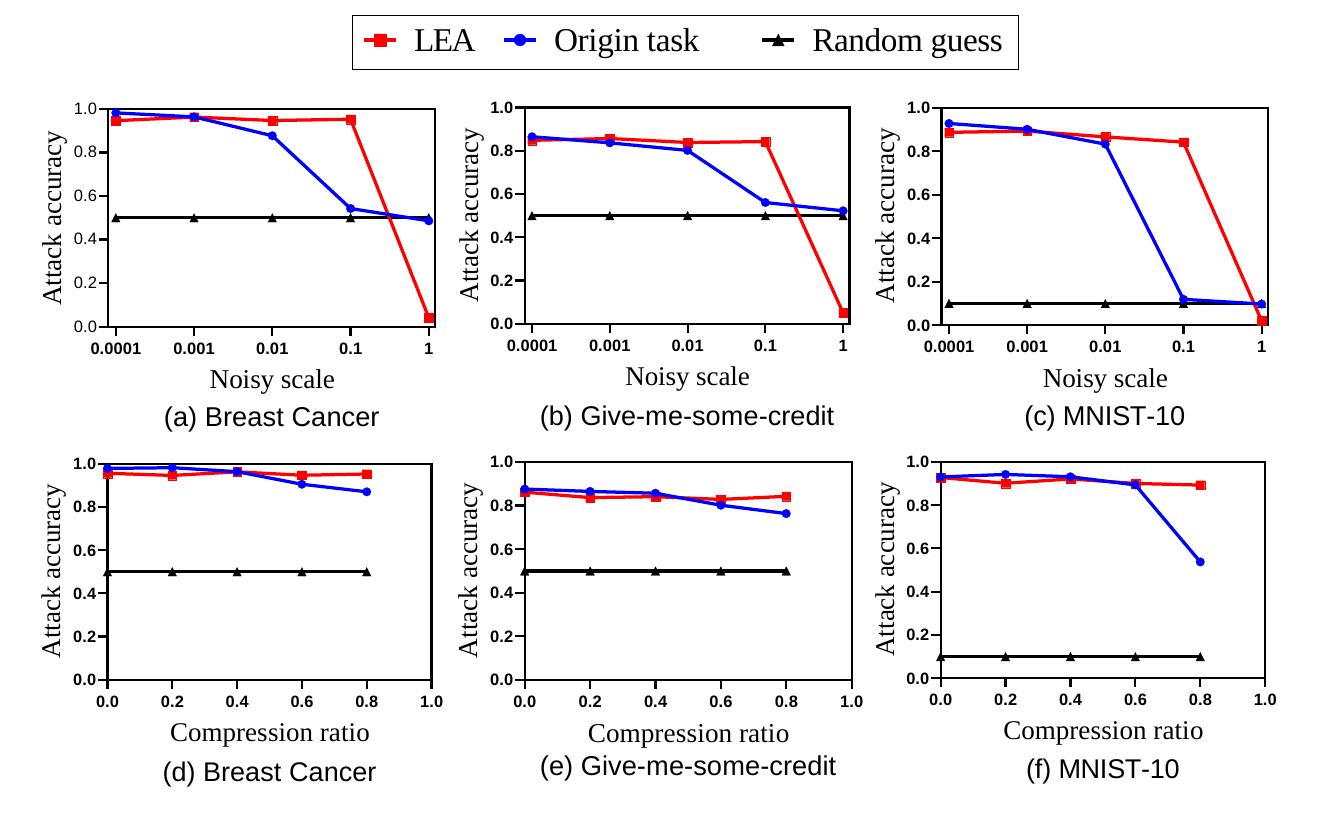}
  \caption{Two defenses against LEA and their impacts. (a)-(c): Noisy Gradients, (d)-(f) Gradient Compression.}\label{fig:protect}
\end{figure}

As depicted in Figure~\ref{fig:protect} (a)-(c), adding Laplace noise of varying magnitudes to the loss gradients shows that even with a significant drop in original task accuracy at a noise level of 0.1, LEA still maintains high accuracy. This occurs because the noise does not sufficiently alter the order of magnitude of the first-round loss gradient similarity. Despite the reduced similarity of the true label's attack model, it remains the most similar. Excessive noise, however, can change the gradient order, leading to incorrect model selection and predictions.

In Figure~\ref{fig:protect} (d)-(f), we compressed gradients by different proportions and found that Breast Cancer and Give-me-some-credit datasets, due to their significant features that can largely determine labels, are not sensitive to changes in the compression ratio. MNIST dataset, however, showed a decrease in accuracy with a higher compression ratio. LEA is less sensitive to compression, primarily because the compressed gradient information still implicitly contains important feature information, and the first-round loss gradients of the attack model is more similar compared to the gradients of another model.

To mitigate the privacy threats posed by LEA, we proposes a defense method based on a label mapping table. Before participating in VFL, the active party generates a label mapping table that maps each label to a different label. During the training process, the active party can use the pseudo labels obtained after mapping as the sample labels to calculate the loss gradients and send them back to the passive party. In this way, the global model trained predicts samples, and the active party can map the predicted pseudo labels back to their real labels according to the label mapping table, thus completing the prediction process. For the adversary, without the information of the label mapping table, even if they obtain an attack model, they cannot find the correspondence between the real labels and can only achieve an accuracy rate of random guessing.

However, this defense method is not effective against LEA in the following two situations: (1) The adversary obtains a small amount of auxiliary dataset. If the adversary has some labeled samples, the labels of the clusters where these samples are located are also likely to be determined, and these can be used in conjunction with the attack model to predict the label values of other samples. (2) The label distribution of the dataset varies greatly. For example, in the Give-me-some-credit dataset, the number of samples with label 1 is only 10\% of the samples with label 0. The adversary can directly assign labels to the two clusters obtained by clustering based on the quantity comparison. In contrast, for datasets with a smaller gap in label distribution, such as the MNIST dataset where each type of sample has a similar proportion, it is impossible to directly determine the labels of each cluster, thus defending against LEA.

\begin{figure}[htbp]
  \centering
  \includegraphics[width=\linewidth]{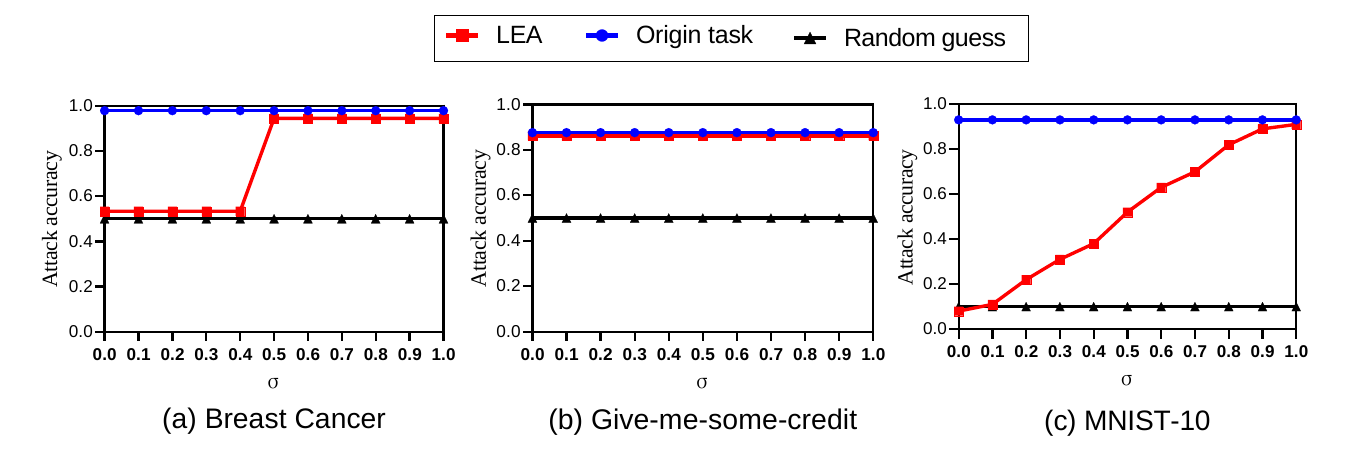}
  \caption{The defense effectiveness of a label mapping table based defense scheme under different proportions of auxiliary datasets owned by adversaries.}\label{defences2}
\end{figure}

To test the effectiveness of the defense strategy based on the label mapping table, let $\sigma$ represent the proportion of the total number of labels that the adversary's auxiliary dataset has. For example, $\sigma=0$ means the adversary has no auxiliary dataset, $\sigma=0.5$ means that the labels of the samples in the adversary's auxiliary dataset account for half of the total labels. In a binary classification task, this means the adversary has samples with one type of label. From Figure \ref{defences2}, it can be seen that the defense strategy based on the label mapping table is effective when the adversary has a small proportion of auxiliary dataset labels. In the binary classification task Breast Cancer dataset, when $\sigma<0.5$, the adversary has no samples with labels, equivalent to having no auxiliary dataset, and under defense it is equivalent to random guessing. However, when $\sigma \geq 0.5$, the adversary has a cluster corresponding to one type of label, which means the label of the other cluster is determined, and the correspondence between the cluster and the label is completely obtained, so it cannot be defended. For the binary classification task Give-me-some-credit dataset, the difference in label distribution is too large, close to 10:1, so the adversary can directly determine the corresponding label based on the size of the cluster obtained by clustering, and LEA attacks cannot be defended. For the MNIST dataset, as $\sigma$ increases, the number of labels the adversary has increases from 1 to 10, and the correspondence between the cluster and the label also becomes clearer. Therefore, there is a positive correlation between the attack accuracy and $\sigma$, which means that when the adversary has many labeled samples, the defense strategy is difficult to defend against LEA attacks.

We evaluated two mainstream defensive methods, including gradient noise and gradient compression. The experimental results indicate that adding noise to the loss gradient to some extent or compressing the gradients does not alter the relative magnitude of the first-round loss gradient similarity. In addition, we propose a defense method based on a label mapping table that can partially defend against LEA, but may fail in specific situations. Therefore, our label enumeration attack reveals the urgency of mitigating label privacy threats in VFL.

\section{Conclusion}

The application of VFL in the industry is increasingly widespread, with privacy and security concerns affecting its implementation. Label information is fundamental and crucial in VFL, and previous work has proposed attack and defense methods for it. We introduce a novel attack method, Label Enumeration Attack (LEA), which infers the label information of the active party using unsupervised clustering from the passive party's data, covering various VFL scenarios. Our experiments show that LEA performs well on real-world datasets. We also propose an efficient Binary-LEA, reducing the computational overhead from  $\mathcal{O}(n!)$  to $\mathcal{O}(n^3)$. We assessed two classic defensive methods, found them ineffective against our attack, and proposed a defense method based on a label mapping table. We hope our work will further drive the exploration of attack methods and defenses in VFL.

\section*{Acknowledgment}
This work was supported by National Natural Science Foundation of China No.62472431.

\bibliographystyle{IEEEtran}
\bibliography{sample-base}

\end{document}